\documentclass{article}

\usepackage{arxiv}

\usepackage[utf8]{inputenc}
\usepackage{lmodern}
\usepackage{microtype}
\usepackage{graphicx}
\usepackage{booktabs}
\usepackage{tabularx}
\usepackage{array}
\usepackage{amsmath,amssymb}
\usepackage{enumitem}
\usepackage{cite}
\usepackage{caption}
\usepackage{float}
\usepackage{xurl}
\usepackage[colorlinks=true,linkcolor=blue,citecolor=blue,urlcolor=blue]{hyperref}

\graphicspath{{figures/}}
\title{TRIP-Evaluate: An Open Multimodal Benchmark for Evaluating Large Models in Transportation}

\author{
 Han Gong \\
  School of Transportation\\
  Southeast University\\
  Nanjing, Jiangsu 211189 \\
  \texttt{213240417@seu.edu.cn} \\
   \And
 Zhen Zhou \\
  School of Transportation\\
  Southeast University\\
  Nanjing, Jiangsu 211189 \\
  \texttt{zzhou602@seu.edu.cn} \\
  \And
 Yunyang Shi \\
  School of Artificial Intelligence and Computer Science\\
  Jiangnan University\\
  Wuxi, Jiangsu 214122 \\
  \texttt{yul217@pitt.edu} \\
  \And
 Yan Tan \\
  School of Transportation\\
  Southeast University\\
  Nanjing, Jiangsu 211189 \\
  \texttt{3228735639@qq.com} \\
  \And
 Jinbiao Huo \\
  Department of Civil and Environmental Engineering\\
  Hong Kong Polytechnic University\\
  Kowloon, Hong Kong \\
  \texttt{jinbiao.huo@polyu.edu.hk} \\
\And
 Qi Hong \\
  School of Transportation\\
  Southeast University\\
  Nanjing, Jiangsu 211189 \\
  \texttt{hongqi@seu.edu.cn} \\
  \And
 Zhiyuan Liu \\
  School of Transportation\\
  Southeast University\\
  Nanjing, Jiangsu 211189 \\
  \texttt{zhiyuanl@seu.edu.cn} \\
}

\begin{document}
\maketitle

\begin{abstract}
Large language models (LLMs) and multimodal large models (MLLMs) are
increasingly used for transportation tasks such as regulation question
answering, traffic management support, engineering review, and
autonomous-driving scene reasoning. Yet transportation workflows are
rule-intensive, computation-intensive, safety-critical, and inherently
multimodal. Existing general benchmarks provide limited evidence of
whether a model can apply regulations correctly, perform verifiable
engineering calculations, or interpret traffic scenes reliably, while
the small number of public transportation benchmarks remain narrow in
scope and rarely support fine-grained diagnosis across text, images, and
point-cloud data. To address this gap, we present TRIP-Evaluate, an open
multimodal benchmark for large models in transportation. The benchmark
organizes 837 items using a role-task-knowledge taxonomy that covers
vehicle, traffic-management, traveler, and planning-and-design
functions. Each item is annotated with capability, modality, and
difficulty labels, enabling diagnosis from overall accuracy down to
specific failure modes. The current release includes 596 text items, 198
image items, and 43 point-cloud items. TRIP-Evaluate also standardizes
item construction, quality control, prompting, decoding, and scoring to
improve cross-model comparability. Results on a diverse panel of models
show that text-based performance is improving, but substantial
weaknesses remain in multi-step engineering calculation,
rule-constrained reasoning, multimodal scene understanding, and
point-cloud understanding. Overall, TRIP-Evaluate provides a
reproducible, diagnosable, and engineering-aligned evaluation baseline
for model selection, regression testing, and safer deployment in
transportation applications.
\end{abstract}


\section{Introduction}\label{introduction}

\subsection{Background and Motivation}\label{background-and-motivation}

As foundation models continue to improve, researchers and practitioners
are exploring their use in transportation tasks such as regulatory
question answering, traffic control support, engineering review, and
autonomous-driving scene reasoning \cite{ref1,ref2}. Transportation tasks,
however, impose requirements that differ from those of open-domain
conversation. They are rule-intensive because outputs must remain
consistent with regulations, standards, and operating procedures. They
are computation-intensive because many tasks require unit conversion,
engineering formulas, and boundary-condition checks. They are also
safety-critical because a plausible but incorrect answer can lead to
compliance failures, design errors, or unsafe operational advice.

Transportation problems are further complicated by their multimodal
nature. A usable model must read statutes, standards, and technical
documents, but it must also interpret road images, signs, markings,
intersection geometry, and sometimes three-dimensional point clouds. In
real workflows, these inputs are not optional. They often determine
whether a rule applies, whether a calculation is valid, and whether an
action is safe in context. A model that performs well on generic
benchmarks can therefore still fail in transportation practice.

A transportation benchmark should test at least three properties. First,
it should test rule consistency, including preconditions, exceptions,
and priority relations. Second, it should test numerical validity, so
that engineering calculations remain reproducible and dimensionally
consistent. Third, it should test contextual grounding, so that the
model can connect a decision to the specific scene in which it is made.
Together, these requirements motivate a benchmark organized around real
transportation work rather than generic knowledge categories.

\subsection{Gaps in Existing Benchmarks}\label{gaps-in-existing-benchmarks}

Most mainstream benchmarks for large models emphasize general knowledge,
broad reasoning, or mixed academic tasks. Although they are useful for
measuring foundational capabilities, they rarely capture the
professional constraints that shape real transportation practice. As a
result, a model can achieve a strong aggregate score yet still be unfit
for real transportation work.

Current benchmarks fall short in four ways. First, domain coverage
remains limited: general benchmarks rarely probe traffic regulations,
engineering standards, parameter verification, signal timing, or safety
auditing in sufficient depth. Second, diagnosis remains limited:
aggregate scores do not reveal which role, task, or knowledge point is
responsible for failure. Third, modality support remains limited:
transportation reasoning depends heavily on visual and spatial evidence,
so text-only evaluation cannot adequately test signs and markings,
intersection layouts, or channelization geometry. Fourth,
reproducibility remains limited: differences in prompts, output formats,
sampling settings, and scoring rules can introduce substantial
evaluation noise.

Existing transportation benchmarks also tend to concentrate on narrow
subtasks such as perception, prediction, and local decision-making in
autonomous driving. Many are built around overseas data and rule
systems, which limits their alignment with domestic regulations,
infrastructure, and operational workflows. As a result, they struggle to
cover the full transportation decision chain from regulations and
engineering to safety and management, making it easier for models to
achieve respectable scores without being genuinely useful in practice.

\subsection{Objective and Contributions}\label{objective-and-contributions}

To address these limitations, we propose TRIP-Evaluate, an open
multimodal benchmark for large models in transportation. Its goal is to
provide fine-grained, interpretable diagnostic signals while preserving
automated scoring and cross-model comparability, thereby supporting
model iteration and deployment risk assessment in transportation.

This paper makes three main contributions.

First, it introduces a transportation benchmark organized around a
role-task-knowledge taxonomy spanning four major functions: vehicle
operations, traffic management, traveler services, and
planning-and-design functions.

Second, it assigns multidimensional labels for capability, modality, and
difficulty, enabling results to be decomposed from overall performance
into business-oriented and technically interpretable slices.

Third, it standardizes benchmark construction and evaluation through
quality-control rules, unified prompting and scoring, and a
deterministic point-cloud rendering pipeline, thereby reducing
interface-driven noise and improving reproducibility.

\section{Related Work}\label{related-work}

\subsection{General Benchmarks and Domain Evaluation}\label{general-benchmarks-and-domain-evaluation}

General-purpose benchmark suites typically emphasize broad knowledge,
reading comprehension, mathematical reasoning, and mixed-task
performance \cite{ref3,ref4,ref5,ref21}. They are valuable for comparing foundational
capabilities, but their tasks and constraints are rarely aligned with
the rule systems and engineering workflows of a specific domain. In
transportation, the most consequential failures often arise from the
incorrect application of regulations, omitted boundary conditions, or
unsupported recommendations in safety-critical settings.

TRIP-Evaluate addresses this gap by treating transportation evaluation
as a diagnostic problem, using a three-level taxonomy together with
capability, modality, and difficulty labels. Recent work on testing and
evaluation in autonomous driving similarly highlights the need for more
targeted validation under realistic operational constraints \cite{ref22}.

\subsection{Transportation Data Resources}\label{transportation-data-resources}

Transportation already has many public datasets and benchmarks, but most
were not designed to evaluate language-centered or multimodal reasoning
in large models. One major group consists of autonomous-driving
perception and scene-understanding datasets such as nuScenes, the Waymo
Open Dataset, DAIR-V2X, sign benchmarks, lane benchmarks, and related
evaluation resources \cite{ref6,ref7,ref8,ref9,ref10,ref11}. These datasets provide realistic scenes
and objective metrics for perception, tracking, and lane understanding,
but they do not test whether a model can correctly apply regulations,
engineering standards, or decision constraints.

A second group consists of trajectory benchmarks such as T-Drive, highD,
NGSIM, the US Highway 101 dataset, pNEUMA, and Argoverse 2 \cite{ref12,ref13,ref14,ref15,ref16,ref17}.
These are valuable for behavior modeling and interaction forecasting,
yet their labels are centered on trajectories rather than on
rule-consistent reasoning or explainable decision making. A third group
consists of traffic forecasting and spatiotemporal prediction resources,
including operational data sources such as PeMS and city-scale
prediction benchmarks \cite{ref18,ref19,ref20}. These datasets are operationally
realistic, but they are dominated by structured temporal signals and
rarely connect scene semantics, regulatory constraints, and multi-step
reasoning.

\subsection{Transportation Benchmarks for Large Models}\label{transportation-benchmarks-for-large-models}

A small number of recent benchmarks move closer to the intersection of
language and transportation, including TransportationGames, DriveLM,
SUTD-TrafficQA, Talk2Car, and explanation-oriented self-driving datasets
\cite{ref23,ref24,ref25,ref26,ref27,ref28}. These efforts are important because they introduce
natural-language supervision and cross-modal alignment into the
transportation domain.

However, they still focus primarily on autonomous-driving scenarios,
visual question answering, reference resolution, or explanation
generation. They do not yet provide broad coverage of transportation
regulation, engineering calculation verification, and strongly
constrained safety reasoning within a unified diagnostic framework.
Because transportation tasks are inherently multimodal, a benchmark for
large models in transportation should support text, images, and point
clouds from the outset and should make cross-modal comparison an
explicit part of the protocol rather than an afterthought.

\section{Methods}\label{methods}

\subsection{Design Goals and Benchmark Scope}\label{design-goals-and-benchmark-scope}

TRIP-Evaluate is designed around three principles: comparability,
diagnosability, and extensibility. Comparability requires standardized
question format, prompting, decoding, and scoring so that interface and
inference-setting differences introduce as little noise as possible.
Diagnosability requires a hierarchical taxonomy and multidimensional
labels so that the benchmark reports interpretable profiles rather than
only a single score. Extensibility requires that new knowledge points
and multimodal items can be added without changing the
benchmark's organizing logic.

The benchmark is not intended only to rank models. It is designed to
reveal whether a model can perform transportation calculations reliably,
whether its errors cluster around units or boundary conditions, and
whether its reasoning process can be audited when needed. For example, a
stopping-sight-distance item can distinguish among incorrect speed
conversion, omission of reaction distance, and misuse of a downgrade
term. These distinctions become visible when results are grouped by
capability and difficulty rather than reported only as overall accuracy.

\subsection{Dataset Composition and Item Format}\label{dataset-composition-and-item-format}

The current TRIP-Evaluate core test set contains 837 items and is
intentionally multimodal. It includes 596 text items, 198 image items,
and 43 point-cloud items. The text subset focuses on regulations,
standards, engineering calculations, and operational judgment. The image
subset covers visual traffic evidence such as signs, markings,
intersection layouts, and channelization. The point-cloud subset
provides an initial probe of three-dimensional structure understanding
and spatial-relation reasoning.

All items are currently formulated as single-choice questions with one
correct answer and four options (A/B/C/D). Text items are presented as a
question with options. Image items are presented as an image, a
question, and options. Point-cloud items are presented as two
deterministic renderings, a bird's-eye view (BEV) image and a front-view
image, together with the question and options. Each item also retains an
explanation field for human audit and future extension to
explanation-based evaluation.

\begin{table}[t]
\centering
\caption{Core Item Schema Used in TRIP-Evaluate}
\label{tab:item-schema}
\small
\begin{tabularx}{\textwidth}{>{\raggedright\arraybackslash}p{0.18\textwidth} >{\raggedright\arraybackslash}p{0.12\textwidth} X}
\toprule
\textbf{Field} & \textbf{Type} & \textbf{Description} \\
\midrule
\texttt{id} & String & Unique sample identifier (e.g., \texttt{B\_b\_1}) \\
\texttt{role\_tag} & String & First-level category: vehicle role / traffic management role / traveler role / planning and design role \\
\texttt{subject\_tag} & String & Second-level category: specific task scenario (e.g., traffic safety) \\
\texttt{knowledge\_point} & String & Third-level category: specific knowledge point (e.g., sight triangle) \\
\texttt{question} & String & Question stem, including the scenario setup and the problem statement \\
\texttt{A}, \texttt{B}, \texttt{C}, \texttt{D} & String & Text content of the answer options (corresponding to the four candidate choices) \\
\texttt{answer} & String & Correct answer option \\
\texttt{explanation} & String & Answer explanation, citing regulatory provisions or calculation formulas \\
\texttt{difficulty} & String & Difficulty level: easy / medium / hard \\
\texttt{modality} & String & Data modality: text / image / point cloud \\
\texttt{capability\_tag} & String & Capability dimension: knowledge memory / logical reasoning / numerical calculation / scene understanding \\
\bottomrule
\end{tabularx}
\end{table}

\subsection{Role-Task-Knowledge Taxonomy}\label{role-task-knowledge-taxonomy}

The first level of the role-task-knowledge taxonomy is role. Rather than
classifying items only by subject matter, TRIP-Evaluate begins from
transportation functions so that results map directly to real workflows
and responsibility boundaries. The benchmark currently defines four
high-level roles.

\begin{enumerate}[label=\arabic*.]
\item
  Vehicle role: perception, localization, planning, control, and driving
  safety.
\item
  Traffic management role: signal control, operations monitoring,
  incident response, facility maintenance, and enforcement compliance.
\item
  Traveler role: trip planning, travel decision making, reliability, and
  information consultation.
\item
  Planning and design role: geometric design review, demand forecasting,
  capacity evaluation, and traffic safety audit.
\end{enumerate}

Under each role, TRIP-Evaluate defines a second-level task domain and a
third-level knowledge point. The current core benchmark covers 16 task
domains and 226 knowledge points. The newly expanded image items add
two knowledge points under existing task domains, so the number of task
domains remains unchanged. This structure turns the benchmark
into an enumerable, extensible, and diagnosable evaluation space rather
than a loose collection of examples.

\begin{figure}[H]
\centering
\includegraphics[width=\linewidth]{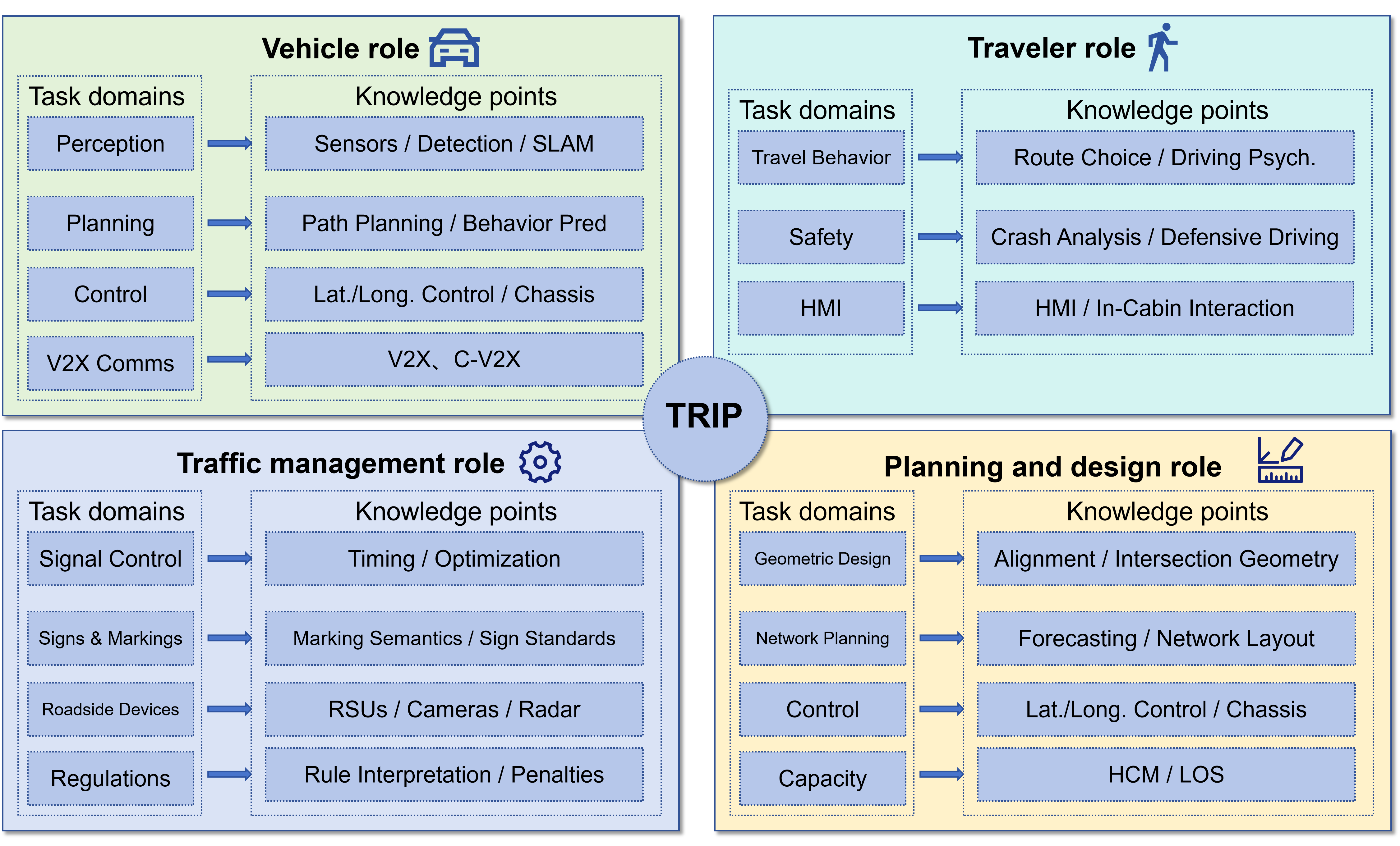}
\caption{Example of the three-level TRIP-Evaluate taxonomy}
\label{fig:taxonomy}
\end{figure}

\subsection{Annotation Dimensions}\label{annotation-dimensions}

Beyond the role-task-knowledge taxonomy, each item is labeled by
capability, modality, and difficulty.

The capability dimension includes four values. Knowledge memory captures
direct recall of regulations, terminology, and standard parameter
ranges. Logical reasoning covers multi-step conditional judgment, causal
reasoning, and conflict resolution among rules. Numerical calculation
targets engineering calculation, unit conversion, and uniquely
verifiable numerical inference. Scene semantic understanding captures
structures, relationships, and behavioral intent in specific traffic
scenes, including image and point-cloud inputs.

The modality label indicates whether the item is Text-only, Image, or
Point Cloud. An item is labeled as multimodal only when the visual
carrier is necessary for solving it; decorative visuals are excluded to
avoid artificial evaluation noise. The difficulty label has three
levels, easy, medium, and hard, based on reasoning depth,
boundary-condition complexity, and distractor strength.

\subsection{Data Construction and Quality Control}\label{data-construction-and-quality-control}

The construction pipeline is organized to support reliable scoring,
reproducibility, and cross-model comparability. The current core set
contains 171 vehicle items, 281 traffic management items, 191 traveler
items, and 194 planning-and-design items. Construction followed an
iterative, versioned process: early versions emphasized high-consistency
items to validate the full authoring and evaluation pipeline, and later
versions expanded coverage, filled long-tail knowledge points, and
rebalanced roles. At the capability level, the benchmark contains
118 knowledge-memory items, 200 logical-reasoning items, 255
numerical-calculation items, and 264 scene-understanding items. By
difficulty, it contains 167 easy items, 456 medium items, and 214
hard items.

Four quality-control rules are central. First, determinacy and
repeatability are required. Every item must have a unique, auditable
answer. Regulation items specify the relevant clause or standard basis,
and engineering items record formulas, variables, units, and required
intermediate quantities. Second, distractor quality is enforced through
a near-miss strategy in which each distractor introduces one
identifiable error, such as a missing condition, a boundary mistake, or
a confused concept.

Third, the benchmark explicitly addresses \textbf{length bias} in
multiple-choice evaluation \cite{ref30}. If
\(l_A\),\(l_B\),
\(l_C\), and \(l_D\) denote nonblank
option lengths, the benchmark constrains

\[
 r = \frac{\max(l_A,l_B,l_C,l_D)}{\min(l_A,l_B,l_C,l_D)}.
\]

The hard threshold is \(r \leq 1.25\). A limited soft-pass window
of \(1.25 < r \leq 1.35\) is allowed only when the correct
option is not the longest and the shortest option is not obviously
perfunctory. Fourth, \textbf{answer distribution and version
consistency} are monitored through approximate A/B/C/D balance,
duplicate screening, schema checks, random recalculation, and regression
repair \cite{ref31,ref32}.

\subsection{Deterministic Point-Cloud Rendering}\label{deterministic-point-cloud-rendering}

Most visual large-language models are pretrained on two-dimensional
images, whereas raw point clouds are sparse and unordered. To reduce
this representation gap without relying on opaque feature extractors,
TRIP-Evaluate converts point clouds into two deterministic views: a
\textbf{bird's-eye view (BEV)} and a \textbf{front view} \cite{ref33}.
Height or depth encoding, occlusion ordering, and distance attenuation
are used to preserve geometric information in the image domain.

After loading point clouds from formats such as .bin, .pcd, or .npy, the
benchmark assumes the common right-handed autonomous-driving coordinate
system with X forward, Y left, and Z up. For consistent image rendering,
the point cloud is transformed to an internal target system with X
right, Y forward, and Z up:

\[
 X_t = -Y_s, \qquad Y_t = X_s, \qquad Z_t = Z_s.
\]

For BEV rendering, the effective region is cropped to \(X \in [-40,40]\) m, \(Y \in [0,80]\) m, and \(Z \in [-2,4]\) m, then rasterized at 0.1 m/pixel into an \(800 \times 800\) image. For front-view rendering, the benchmark
uses a virtual pinhole camera with \(h_{\mathrm{cam}} = 1.5\) m, horizontal and
vertical fields of view of \(90^\circ\) and \(60^\circ\), and an output resolution
of \(800 \times 600\). A point \((x,y,z)\) is projected as

\[
 u = f_x \frac{x}{y} + c_x, \qquad
 v = f_y \frac{h_{\mathrm{cam}} - z}{y} + c_y.
\]

Near-field singularities and overly distant points are removed, and
rendering is ordered from near to far so that closer objects occlude
farther ones. Figure 2 summarizes the pipeline.

\begin{figure}[H]
\centering
\includegraphics[width=\linewidth]{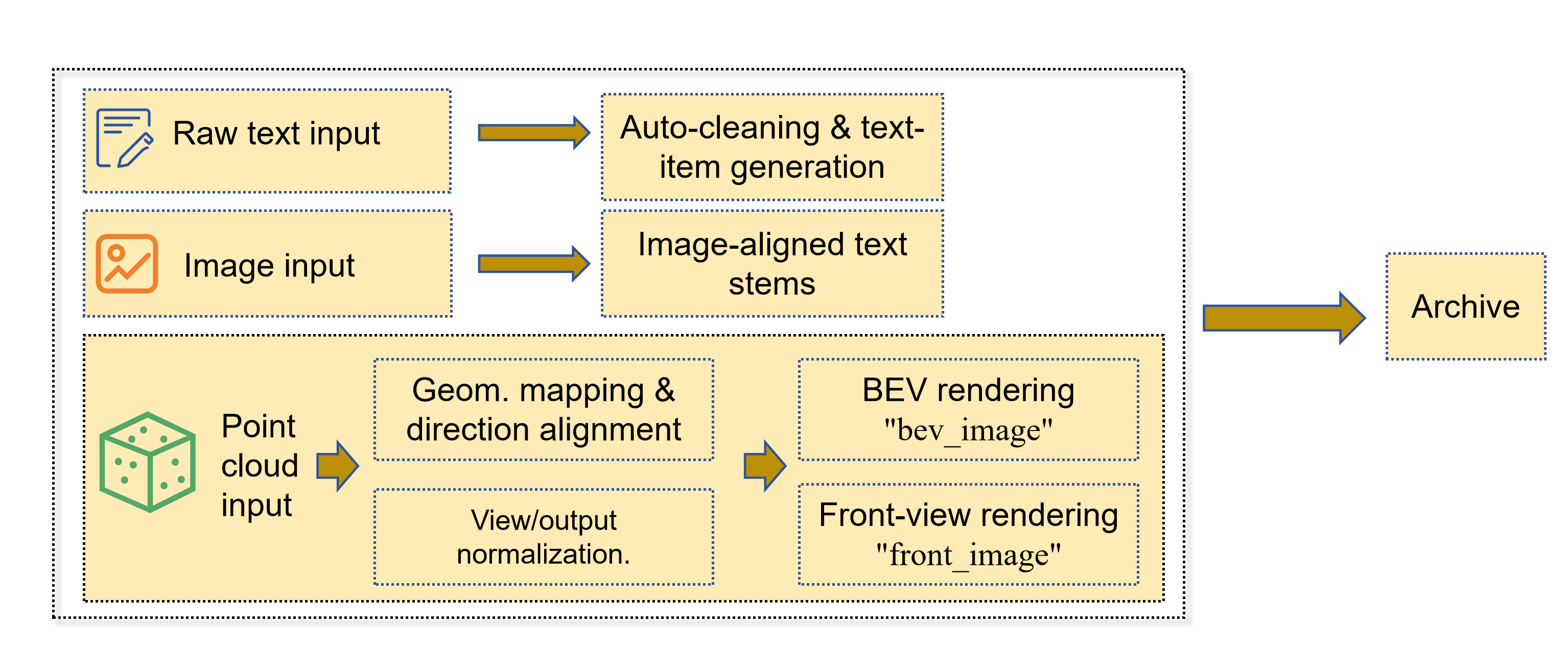}
\caption{Sample generation and processing pipeline in TRIP-Evaluate}
\label{fig:generation-pipeline}
\end{figure}

\subsection{Evaluation Protocol and Metrics}\label{evaluation-protocol-and-metrics}

TRIP-Evaluate adopts a unified question format, output requirement, and
scoring rule. Every item is presented as a single-choice question with
four options. Depending on the modality, the input contains the question
plus options and, when needed, an image or the pair of point-cloud
renderings. The prompt constrains the model to output only one letter
from A/B/C/D.

The same core prompt is used for text, image, and point-cloud items,
with only minimal changes to indicate attached visuals. Decoding and
sampling settings are fixed for reproducibility. Abnormal outputs, such
as empty answers, non-A/B/C/D strings, or multiple characters, are
handled by a unified rule set, either by marking them incorrect under
the protocol or by extracting the first valid option when the rule
allows it, and are logged as a stability indicator.

The primary metric is \textbf{accuracy}:

\[
 \mathrm{Acc} = \frac{1}{N}\sum_{i=1}^{N}\mathbb{I}\!\left(\hat{y}_i = y_i\right).
\]

In addition to overall accuracy, the benchmark reports grouped accuracy
by role\_tag, subject\_tag, knowledge\_point, capability\_tag,
difficulty, and modality. Evaluation outputs also record benchmark and
prompt versions, decoding parameters, and abnormal-output handling rules
so that results remain auditable across releases.

\begin{figure}[H]
\centering
\includegraphics[width=\linewidth]{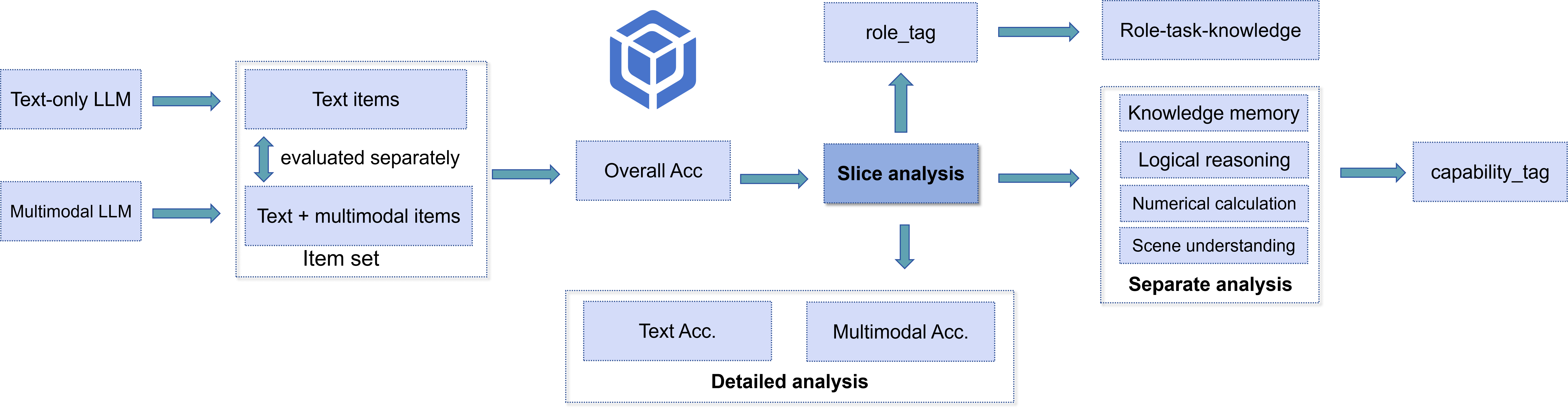}
\caption{Evaluation logic of TRIP-Evaluate}
\label{fig:evaluation-logic}
\end{figure}

\section{Experimental Setup}\label{experimental-setup}

\subsection{Baseline Models}\label{baseline-models}

TRIP-Evaluate evaluates a broad panel of state-of-the-art language and
multimodal models under a unified protocol. The panel spans three
groups. The first group includes reasoning-oriented and proprietary
models, such as DeepSeek-R1 \cite{ref34}, Gemini-3-flash-preview \cite{ref35},
Claude Sonnet 4.6 \cite{ref36}, Claude Sonnet 4.5 \cite{ref37}, and Qwen-max
\cite{ref38}. The second group includes open-weight or broadly available
multimodal models, such as Qwen2-VL-72B-Instruct \cite{ref39},
Qwen3-VL-8B-Instruct \cite{ref40}, gpt-oss-120b and gpt-oss-20b \cite{ref41},
and the Llama-3.2 vision instruction-tuned series \cite{ref42,ref43}. The third
group includes text-only and specialized models, such as DeepSeek-V3.2
\cite{ref44}, Gemma-2-27b-it, Gemma-2-9b-it \cite{ref45}, Qwen3-8B \cite{ref46}, and
Qwen2.5-Coder instruction models \cite{ref47}.

This mix of reasoning-oriented, multimodal, and text-dominant systems
allows the benchmark to probe both overall performance and the specific
penalties associated with modality expansion and long-chain reasoning.

\subsection{Validity Check Through Parameter Scaling}\label{validity-check-through-parameter-scaling}

A useful benchmark should respond consistently to meaningful differences
in model capacity. To verify that TRIP-Evaluate has discriminative
power, the benchmark compares several open model families at different
parameter scales, including Llama-3.2 Vision Instruct \cite{ref42}, Qwen-VL
Instruct \cite{ref48}, Gemma-2 IT \cite{ref45}, and Qwen2.5-Coder Instruct
\cite{ref47}. The comparison is conducted on the full benchmark under the
same question distribution and scoring protocol.

Figure 4 shows the accuracy gains obtained when moving from smaller to
larger models within the same family.

\begin{figure}[H]
\centering
\includegraphics[width=\linewidth]{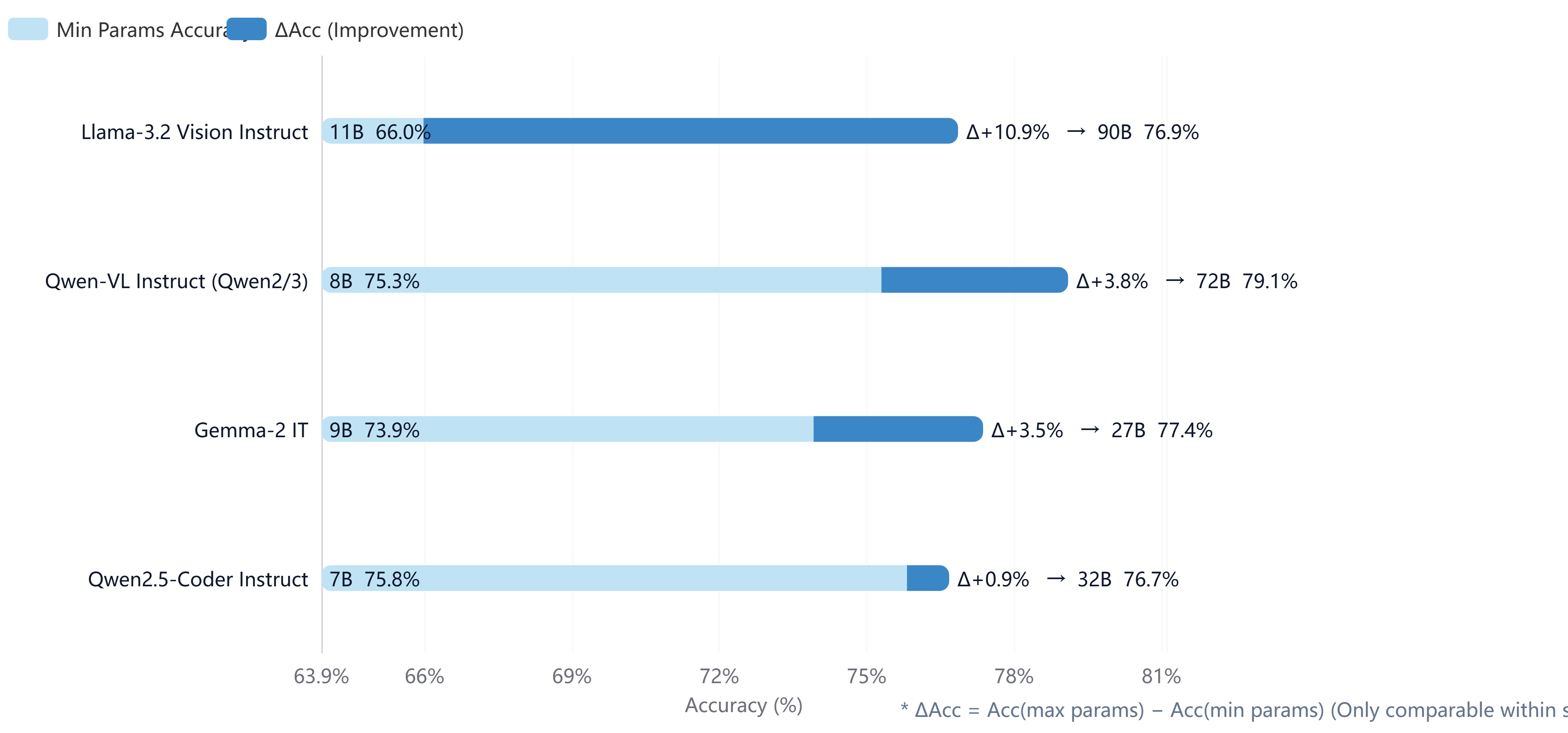}
\caption{Accuracy gains ($\Delta$Acc) within model families across parameter scales}
\label{fig:parameter-scaling}
\end{figure}
Across the families examined, larger models consistently outperform
their smaller counterparts. The gains differ in magnitude but remain
directionally stable, supporting the claim that TRIP-Evaluate captures
meaningful capacity differences rather than random variation.

\section{Results}\label{results}

\subsection{Overall Performance and Model Stratification}\label{overall-performance-and-model-stratification}

TRIP-Evaluate reveals a clear macro pattern: text-side performance is
approaching practical usability, image-side performance is unstable,
point-cloud performance remains inadequate, and hard long-chain tasks
widen the gap among models.

Figure 5 compares overall accuracy on the full multimodal benchmark and
on the text-only subset.

\begin{figure}[H]
\centering
\includegraphics[width=\linewidth]{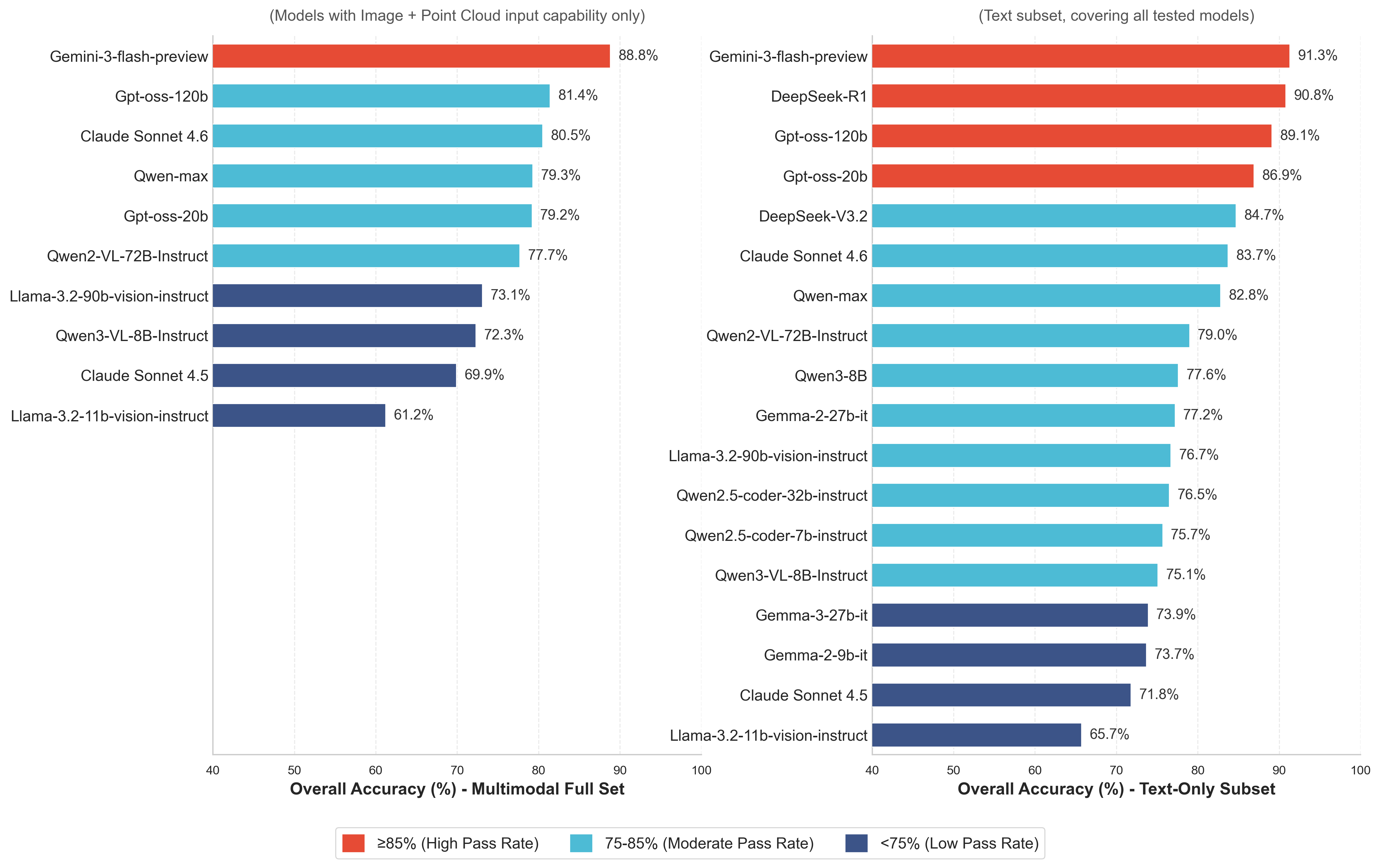}
\caption{Overall accuracy on the full multimodal benchmark and the text-only subset}
\label{fig:overall-accuracy}
\end{figure}
Both settings show a visible ranking structure, but the number of
high-performing models drops on the full multimodal set. On the full
benchmark, only Gemini-3-flash-preview \cite{ref35} reaches or exceeds 85\%
overall accuracy, attaining 88.8\%. Most other models fall between 75\%
and 85\%, and the weakest models remain below 75\%. On the text-only
subset, by contrast, Gemini-3-flash-preview \cite{ref35}, DeepSeek-R1
\cite{ref34}, and the gpt-oss series \cite{ref41} all exceed 85\%, suggesting
that transportation text understanding is improving faster than
multimodal integration.

A paired comparison for models evaluated in both settings shows a strong
positive correlation between text-only and multimodal scores, but also a
stable negative offset. On average, the full multimodal score is about
3.9 percentage points lower than the text-only score, with a median drop
of about 3.35 percentage points. The largest observed drop reaches 7.7
percentage points, as seen for gpt-oss-120b and gpt-oss-20b \cite{ref41}.
Figure 6 visualizes this multimodal retention pattern. Taken together,
the comparison suggests that multimodal input amplifies both performance
dispersion and aggregate uncertainty.

\begin{figure}[H]
\centering
\includegraphics[width=\linewidth]{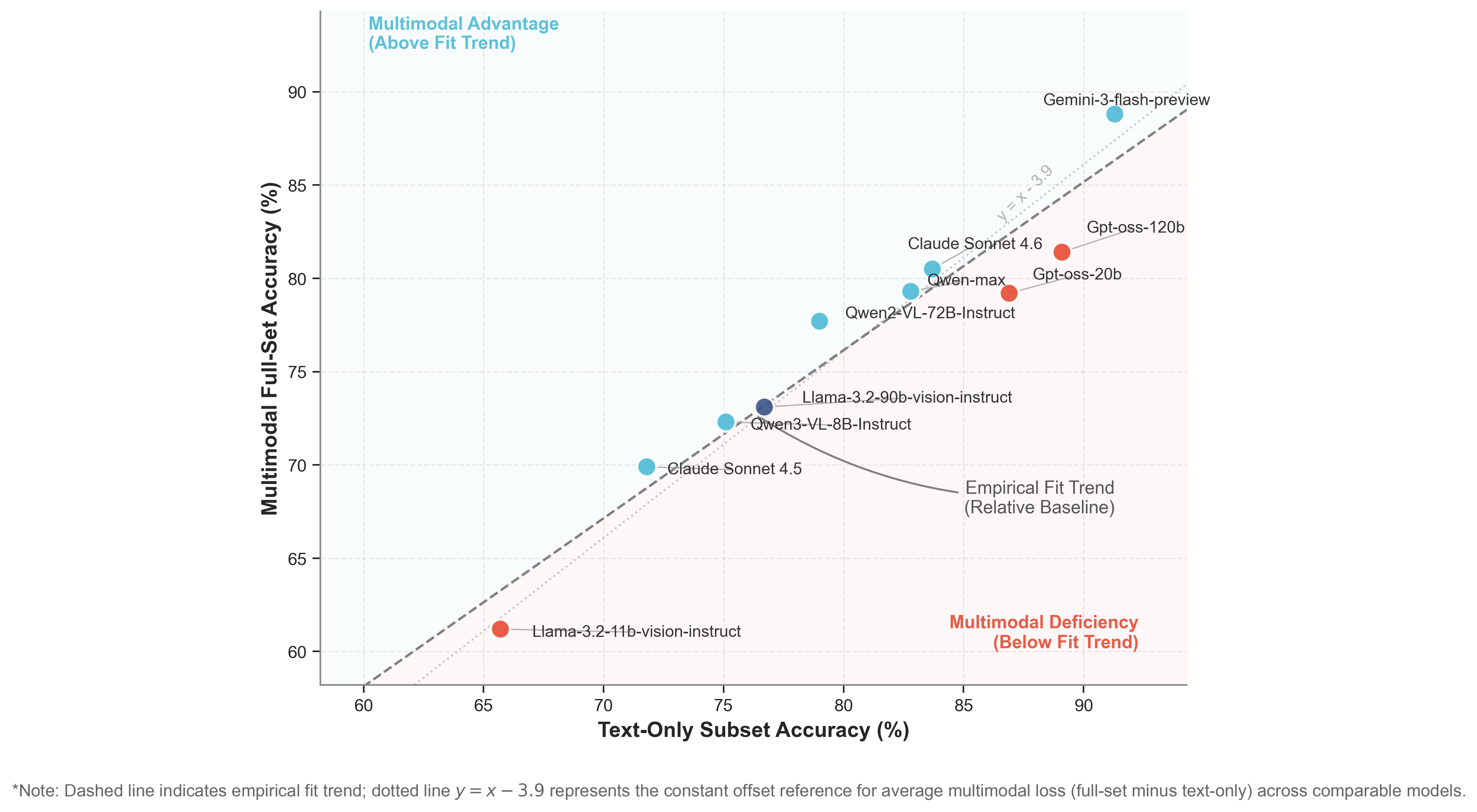}
\caption{Multimodal performance retention relative to text-only accuracy}
\label{fig:multimodal-retention}
\end{figure}
\subsection{Role-Task-Knowledge Diagnosis}\label{role-task-knowledge-diagnosis}

A single aggregate score hides where models succeed and fail. To recover
that structure, TRIP-Evaluate compares performance across the four
transportation roles under both text-only and multimodal conditions.
Figure 7(a) shows role-level accuracy in the text-only setting, and
Figure 7(b) shows the same slicing logic after images and point clouds
are introduced.

\begin{figure}[H]
\centering
\includegraphics[width=\linewidth]{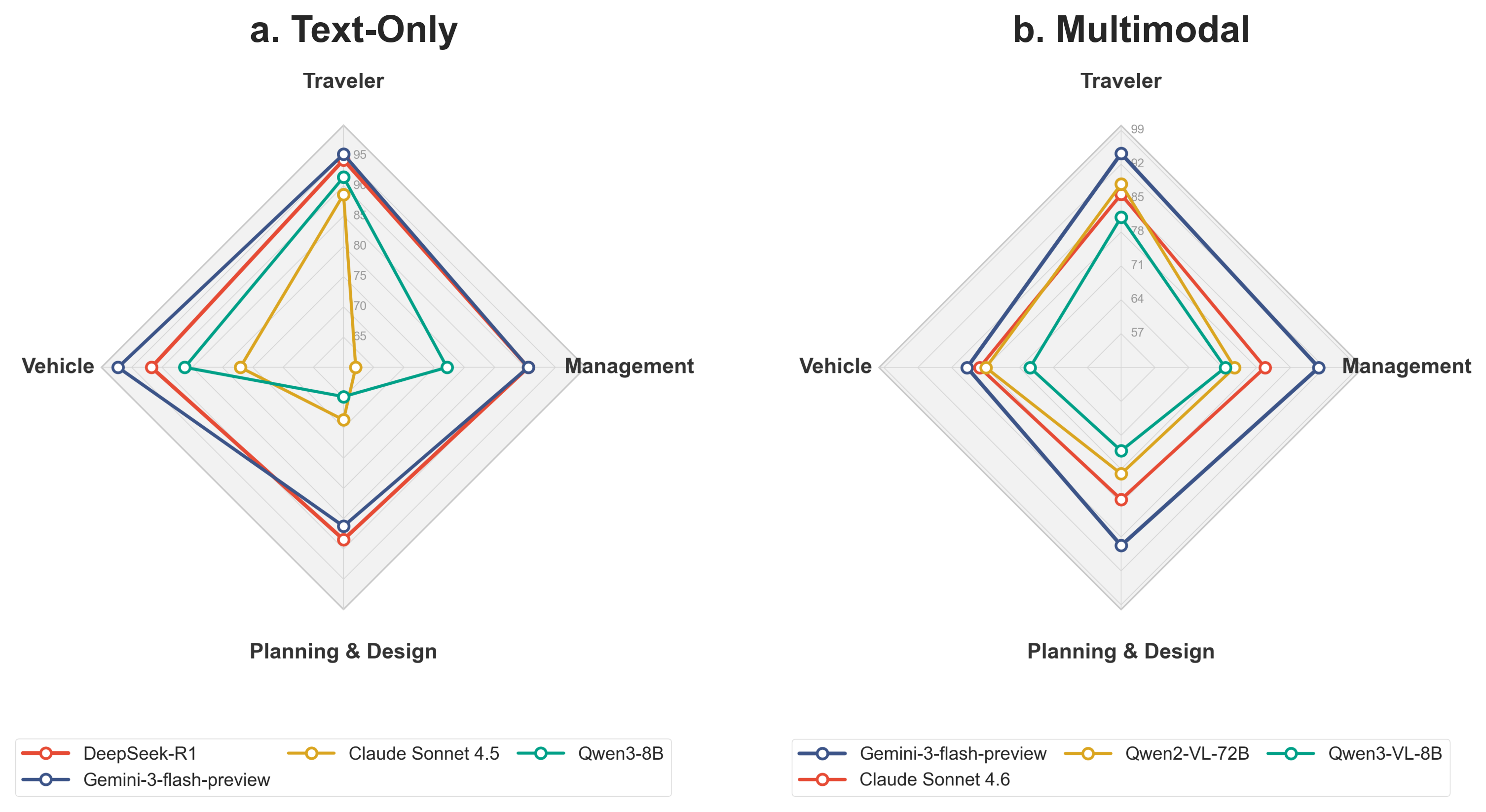}
\caption{Role-level capability profiles under text-only and multimodal settings}
\label{fig:role-profiles}
\end{figure}
The results show a clear imbalance. In text-only settings, leading
models, especially DeepSeek-R1 \cite{ref34}, display relatively broad
coverage across the four roles. Even so, performance is not uniform.
Traveler tasks, which often involve everyday travel decisions or safety
common sense, tend to yield higher accuracy, whereas planning-and-design
tasks, which rely more heavily on engineering standards, formula-based
review, and constraint consistency, are consistently harder.

The multimodal comparison introduces an additional drop, especially on
vehicle tasks that depend on scene perception. For some multimodal
models, including Gemini-3-flash-preview \cite{ref35} and
Qwen2-VL-72B-Instruct \cite{ref39}, role-level performance decreases once
images and point clouds become part of the reasoning chain. This pattern
suggests that cross-modal alignment and three-dimensional spatial
representation remain important bottlenecks.

Figure 8 presents Pareto-style error plots for representative domains
under the four roles.

\begin{figure}[H]
\centering
\includegraphics[width=\linewidth]{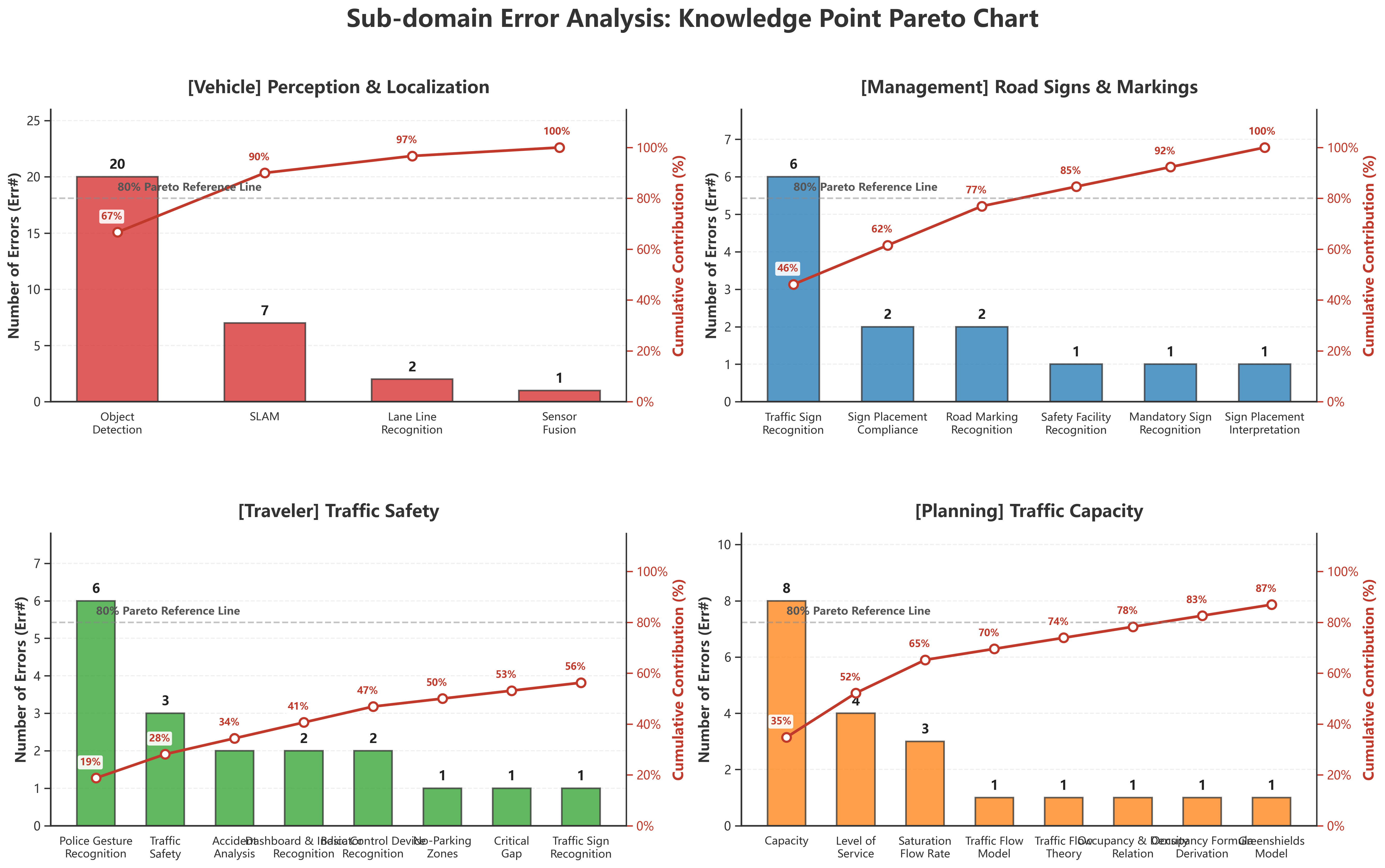}
\caption{Knowledge-point Pareto analysis of errors across the four roles}
\label{fig:pareto-errors}
\end{figure}
Two structural patterns emerge. Errors in the vehicle and traffic
management roles are relatively concentrated in a small set of knowledge
points, whereas traveler and planning-and-design errors are more
distributed. In the vehicle role, object detection alone contributes
about 67\% of the observed errors, and the cumulative contribution
exceeds about 90\% once SLAM is added. In the traffic management role,
traffic-sign-related knowledge points dominate the error profile, and
the cumulative contribution exceeds 80\% only after the fourth knowledge
point, at about 85\%. By contrast, the top eight traveler-side knowledge
points explain only about 56\% of the errors, while the top eight
planning-and-design knowledge points explain about 87\%. These results
indicate that some roles are governed by a few head categories, whereas
others require broader long-tail coverage.

\subsection{Cross-Modal Performance and Modality Penalties}\label{cross-modal-performance-and-modality-penalties}

TRIP-Evaluate next isolates the effect of input modality by comparing
aligned model results across text, image, and point-cloud conditions.
Figure 9 reports accuracy under each modality, and Figure 10 shows the
change relative to the text baseline.

\begin{figure}[H]
\centering
\includegraphics[width=0.92\linewidth]{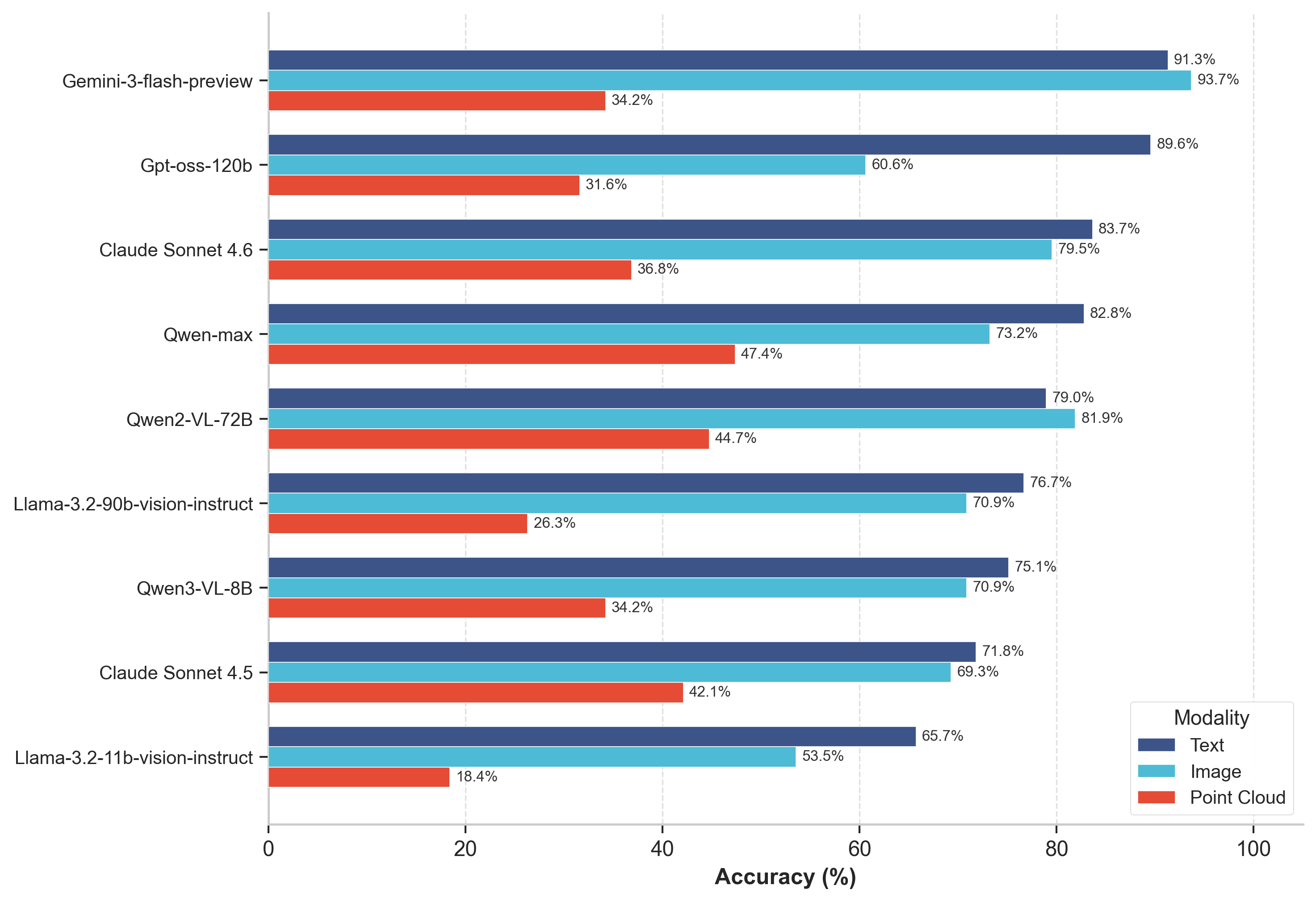}
\caption{Accuracy under text, image, and point-cloud inputs}
\label{fig:modality-accuracy}
\end{figure}

\begin{figure}[H]
\centering
\includegraphics[width=0.92\linewidth]{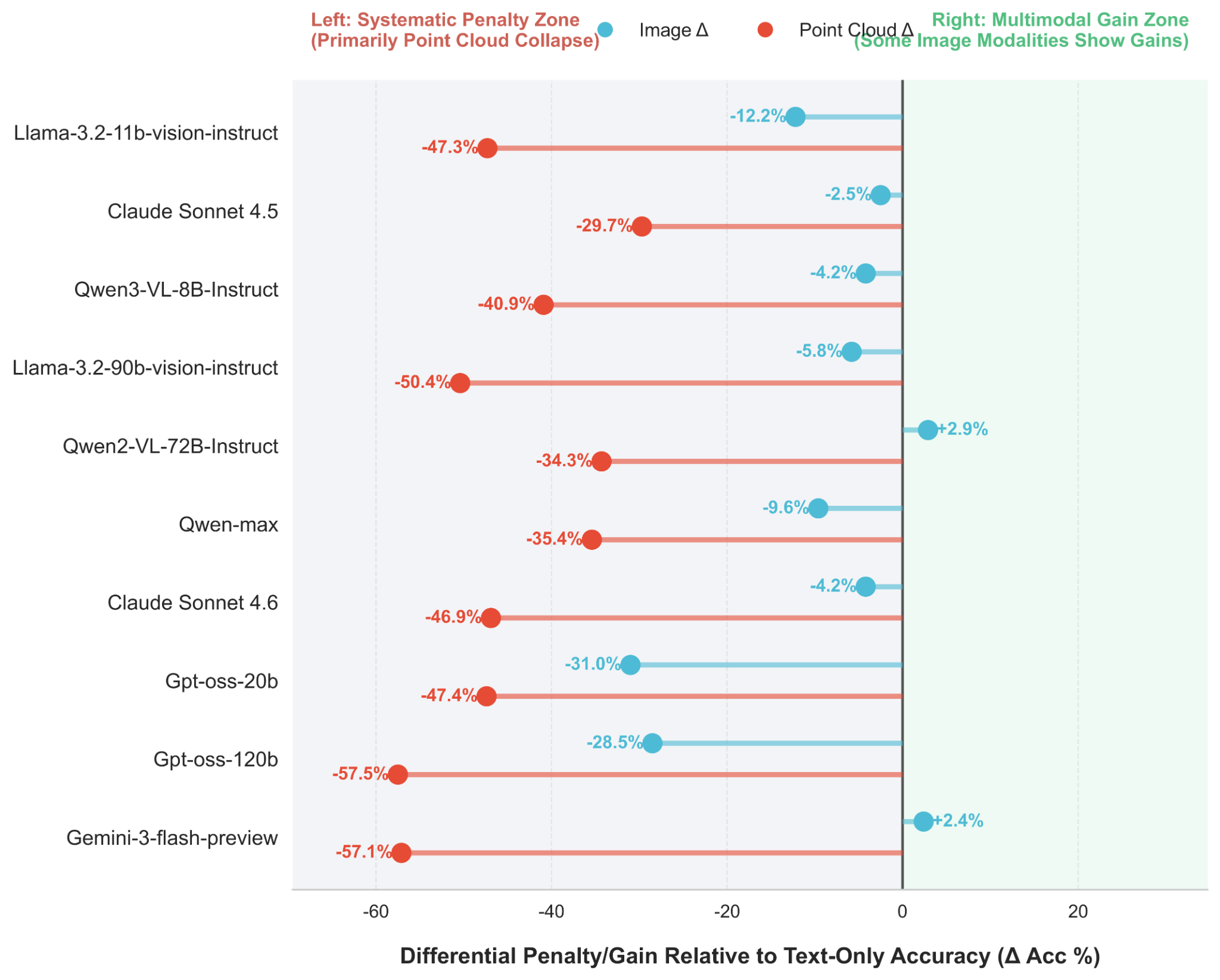}
\caption{Performance change relative to the text baseline ($\Delta$Acc)}
\label{fig:delta-acc}
\end{figure}

Two stable patterns emerge from this comparison. First, image input does
not provide a uniform benefit. A few models gain slightly on image
items, with Qwen2-VL-72B-Instruct and Gemini-3-flash-preview showing
small gains of about 2.9 and 2.4 percentage points, respectively, but
most models decline relative to their text-only results. Second, point
clouds impose a systematic penalty. All evaluated models drop sharply
on point-cloud tasks relative to text, with decreases ranging from
about 29.7 to 57.5 percentage points. This pattern indicates that the
multimodal decline is not simply a matter of harder questions.
More fundamentally, current models still struggle to integrate
heterogeneous evidence into a stable judgment.

The image-side fluctuation likely reflects the dual role of visual input
as both evidence and noise. In some cases, it resolves ambiguity by
exposing lane markings, signs, signal states, or channelization
structure. In others, it introduces blur, glare, occlusion, small-object
uncertainty, or other long-tail disturbances. Point-cloud degradation is
more uniform and therefore more revealing. It points to a deeper
bottleneck in cross-modal alignment and three-dimensional spatial
reasoning: even after deterministic rendering into BEV and front-view
images, the model must still recover topology, relative depth, and
geometric scale from representations that remain only indirectly aligned
with its pretraining distribution.

\subsection{Difficulty Growth and Long-Chain Reasoning}\label{difficulty-growth-and-long-chain-reasoning}

TRIP-Evaluate also separates performance by task difficulty to examine
stability as logical depth increases. As reasoning chains lengthen and
engineering constraints accumulate, some models exhibit what may be
called a pseudo-logic-chain failure mode: the answer is presented as if
it were stepwise and rigorous, but one or more intermediate quantities,
unit checks, or boundary validations are missing or inconsistent.

Figure 11 shows how representative models behave across easy, medium,
and hard items \cite{ref49}.

\begin{figure}[H]
\centering
\includegraphics[width=\linewidth]{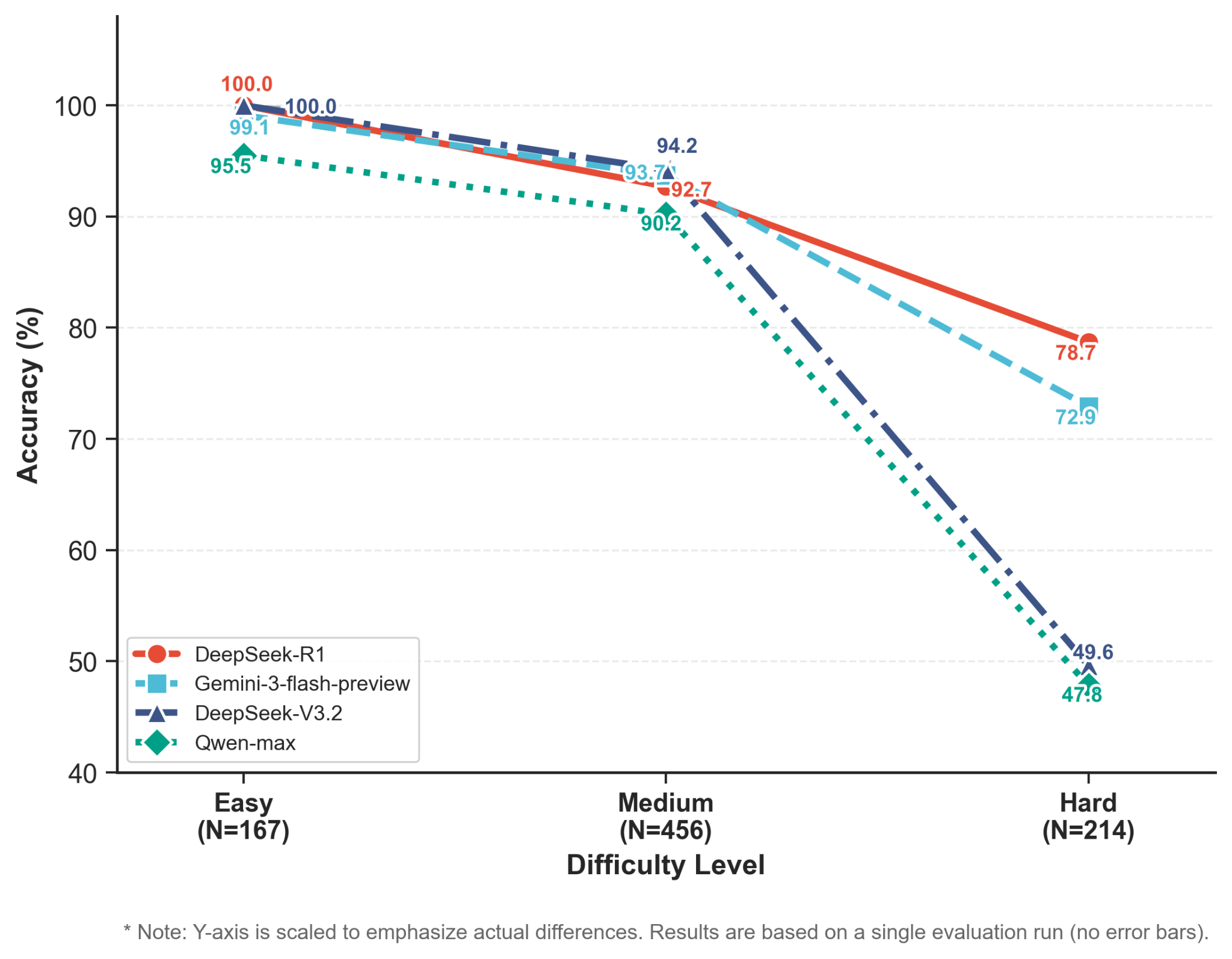}
\caption{Accuracy degradation as difficulty increases}
\label{fig:difficulty-degradation}
\end{figure}

The degradation is not uniform. DeepSeek-V3.2 \cite{ref44}, for example,
remains relatively stable on easy and medium items, but its accuracy
drops sharply on hard items with nested formulas and multiple boundary
conditions, falling to about 49.6\%. Qwen-max \cite{ref38} follows a
similar pattern and reaches about 47.8\% on hard items.
Reasoning-oriented models such as DeepSeek-R1 \cite{ref34} and
Gemini-3-flash-preview \cite{ref35} remain more stable at the hard end,
whereas DeepSeek-V3.2 \cite{ref44} and Qwen-max show much steeper
declines from medium to hard items, at roughly 45 and 42 percentage
points, compared with about 14 and 21 percentage points, respectively,
for the two stronger models.

These results suggest that the gap among models cannot be explained by
differences in knowledge volume alone. Some systems appear better able
to preserve intermediate consistency as additional constraints are
introduced, whereas others degrade more quickly as opportunities for
error propagation accumulate. For high-risk tasks such as engineering
review, capacity calculation, signal timing, and compliance-boundary
judgment, the results support verification-enhanced pipelines that
incorporate deliberation, reranking, or external tools such as
calculators and solvers \cite{ref29,ref49,ref50,ref51}.

\subsection{Capability-Level Analysis Under Dual Pressure Sources}\label{capability-level-analysis-under-dual-pressure-sources}

In addition to business slices, TRIP-Evaluate examines model behavior at
the capability level. From a transportation perspective, two broad
chains are especially important. The first is an engineering-verifiable
chain, which covers formulas, unit consistency, boundary conditions, and
constraint checks. The second is a safety-semantic chain, which covers
scene understanding, right-of-way judgment, rule applicability, and
spatial-relation inference. The benchmark therefore compares capability
slices under two types of pressure: modality switching and difficulty
growth.

Figure 12 summarizes the slice-level results using two structural
penalties: a modality penalty, defined as the gap between text accuracy
and the worst non-text modality, and a cross-difficulty penalty, defined
as the gap between hard and easy conditions.

\begin{figure}[H]
\centering
\includegraphics[width=\linewidth]{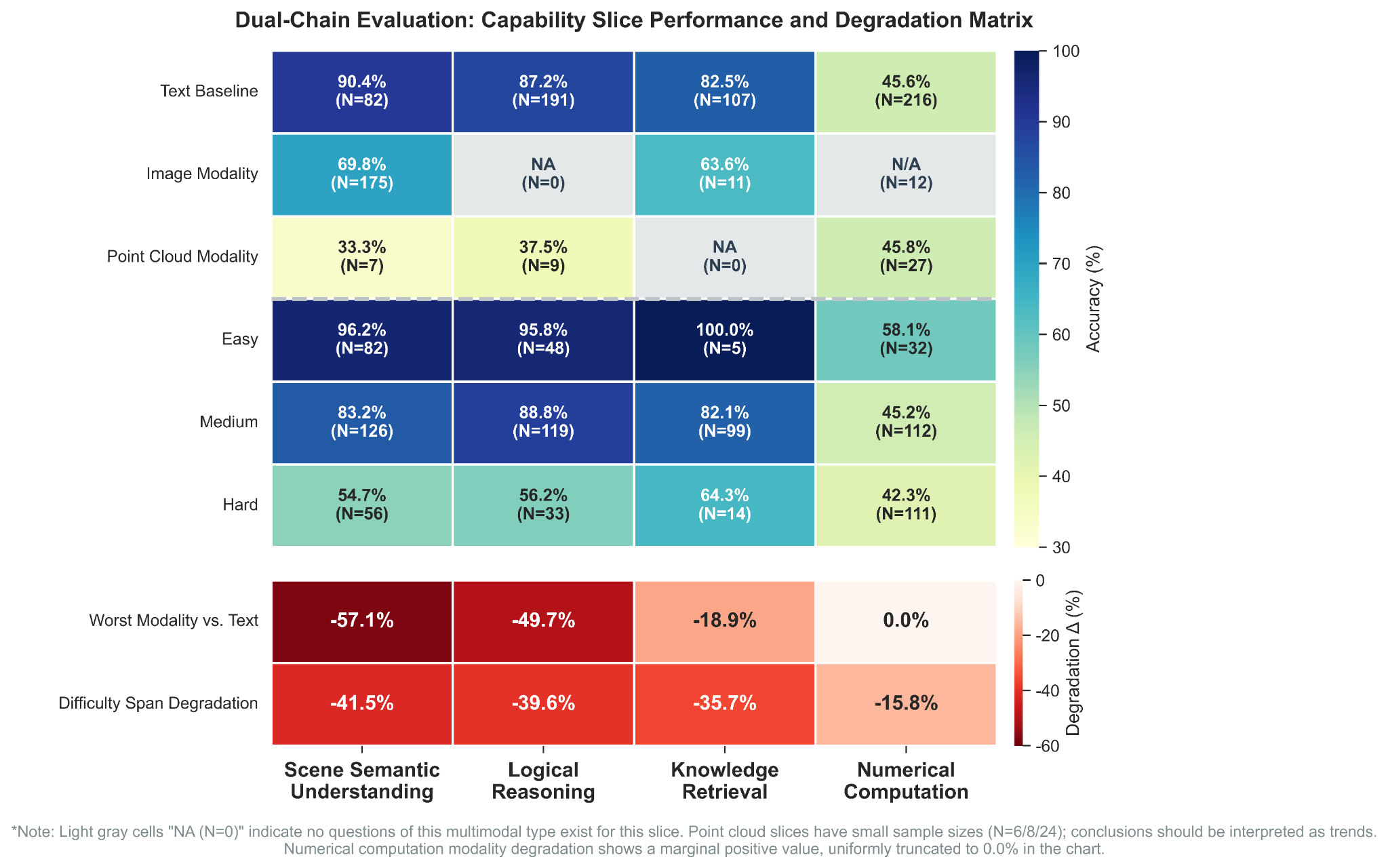}
\caption{Dual-path diagnostic analysis across capability dimensions}
\label{fig:dual-path-diagnostic}
\end{figure}

The results reveal a strong asymmetry. The safety-semantic chain is more
sensitive to modality, whereas the engineering-verifiable chain is more
sensitive to difficulty. Under point-cloud pressure, scene semantic
understanding and logical reasoning show the sharpest declines, dropping
by about 57.1\% and 49.7\%, respectively, relative to the text
condition. Difficulty pressure, by contrast, primarily damages the
engineering-verifiable chain. When the benchmark moves from easy to hard
conditions, scene semantic understanding, logical reasoning, and
knowledge memory decline by about 41.5\%, 39.6\%, and 35.7\%,
respectively.

Numerical calculation appears to be a baseline weakness across
conditions rather than a weakness triggered only by modality. Its
text-side baseline accuracy is only 45.6\%, and its accuracy on hard
items is about 42.3\%. As shown in Figure 12, the harder
engineering-oriented slices still trail the stronger comparison slices
by about 15.8 percentage points. Under multimodal input, the main risk
lies in the safety-semantic chain, especially scene understanding and
spatial reasoning. Under hard problem settings, the main risk lies in
the engineering-verifiable chain, especially long-chain reasoning and
constraint consistency.

\section{Discussion}\label{discussion}

\subsection{What TRIP-Evaluate Reveals About Large Models in Transportation}\label{what-trip-evaluate-reveals-about-large-models-in-transportation}

Taken together, the benchmark results indicate that progress on
transportation tasks remains uneven across current large models. For
text-heavy tasks involving regulations, general transportation
knowledge, and limited reasoning depth, some advanced models may already
be usable under human supervision. Even so, strong text-side performance
can mask structurally important weaknesses.

The most visible gap emerges when tasks shift from pure text to visual
or spatial evidence. Image-based reasoning is not uniformly poor, but it
is unstable: the same visual signal that resolves ambiguity in one case
can introduce noise in another. Point-cloud reasoning remains
substantially harder. The benchmark does not establish a single
architectural cause, but the results strongly suggest that cross-modal
alignment and three-dimensional spatial representation are not yet
dependable enough for broad transportation deployment.

A second gap emerges when tasks require long reasoning chains under
strong constraints. Fluent intermediate text should not be mistaken for
verifiable reasoning. In transportation engineering, correctness depends
on maintaining consistency across formulas, units, and applicability
conditions. This is why the benchmark emphasizes auditable explanations
and slice-level diagnosis rather than final accuracy alone.

\subsection{Practical Implications for Model Selection and Deployment}\label{practical-implications-for-model-selection-and-deployment}

The benchmark has several practical implications for model selection and
deployment. First, model choice should depend on task risk rather than
overall benchmark rank alone. High-risk tasks such as engineering
review, signal-timing verification, and compliance-boundary judgment
should prioritize reasoning-oriented models or systems augmented with
calculators, solvers, or explicit verification modules. Lower-risk tasks
such as traveler information and simple rule lookup may be handled with
lighter architectures and simpler decoding strategies.

Second, improvement strategies should reflect the error structure
revealed by the benchmark. For vehicle and traffic-management tasks,
where errors are concentrated in a few head knowledge points, focused
improvement on those categories may yield disproportionate gains. For
traveler and planning-and-design tasks, where errors are more dispersed,
broader scenario coverage and greater variation in rule expression may
be more effective.

Third, multimodal deployment should be approached cautiously. Adding
images or point clouds does not automatically make a system more
reliable. Additional modalities can create new failure paths, so
multimodal systems should be paired with modality-aware validation,
robustness testing, and, in safety-critical settings, external review.

\subsection{Limitations and Future Work}\label{limitations-and-future-work}

TRIP-Evaluate is an offline benchmark, and offline evaluation
necessarily simplifies real deployment. Transportation regulations and
engineering standards are region-specific and time-sensitive. Real
systems must also cope with distribution shift, noisy inputs, rare edge
cases, and real-time constraints. A static benchmark cannot fully
capture these factors over time or across deployment settings.

Future work can extend the benchmark in at least three directions.
First, retrieval-enhanced or dynamically refreshed evaluation may better
reflect changing regulations and standards. Second, richer temporal
modalities, especially traffic video and sequential point clouds, can
move the benchmark beyond static scenes. Third, closed-loop evaluation
can test not only isolated answers, but also stability and safety
boundaries across longer decision sequences.

\section{Conclusions}\label{conclusions}

This paper presents and open-sources TRIP-Evaluate, a multimodal
benchmark for large models in transportation, to quantify the gap
between general benchmark performance and real engineering usability in
a domain characterized by strong regulatory, computational, and safety
constraints. Organized around a role-task-knowledge taxonomy, the
benchmark supports drill-down diagnosis from overall performance to
task- and knowledge-point-level failure modes. The current release
contains 837 items across text, image, and point-cloud modalities,
including 596 text items, 198 image items, and 43 point-cloud items, and
provides a reproducible basis for cross-model and cross-version
comparison.

The main contributions are twofold. First, TRIP-Evaluate introduces a
hierarchical organization and labeling framework that maps evaluation
results to business responsibility boundaries and enumerable capability
targets, turning a single overall score into interpretable and
actionable diagnostic signals. Second, it provides a deterministic
point-cloud evaluation scheme based on BEV and front-view renderings,
offering a standardized and auditable entry point for three-dimensional
transportation understanding. Together with unified prompting, decoding,
scoring, and quality-control procedures, these design choices reduce
evaluation variance caused by interface and inference-setting
differences.

Slice-level experiments reveal several important capability boundaries
of current large models in transportation. In long-chain engineering
tasks under multiple constraints, instability often appears in
inconsistent units, missing boundary conditions, or unverified
intermediate quantities. Under image and point-cloud inputs, weak
cross-modal alignment and unstable spatial reasoning substantially
amplify uncertainty. For rule-application tasks, complex preconditions,
exceptions, and priority relations remain persistent bottlenecks. Taken
together, these findings suggest a practical deployment principle.
High-risk tasks should prioritize verifiable reasoning pipelines,
whereas lower-risk settings are better served by task-stratified
deployment to balance efficiency and reliability.

Overall, the value of TRIP-Evaluate lies not only in filling the gap in
multimodal transportation evaluation resources, but also in providing a
reproducible, diagnosable, and engineering-aligned evaluation baseline
for model selection, targeted data improvement, regression testing, and
safety-oriented deployment review.

\section{Acknowledgments}\label{acknowledgments}

The authors used ChatGPT (OpenAI) only for English-language polishing
during manuscript preparation. All technical content, results, and final
revisions were reviewed and verified by the authors.

\section{Author Contributions}\label{author-contributions}

The authors confirm contribution to the paper as follows: H. Gong and Z.
Zhou contributed substantially to the study design, benchmark
development, analysis, and manuscript preparation. The other authors
provided support and feedback throughout the research and revision
process. All authors reviewed the results and approved the final version
of the manuscript.

\section{Declaration of Conflicting Interests}\label{declaration-of-conflicting-interests}

The authors declared no potential conflicts of interest with respect to
the research, authorship, and/or publication of this article.

\section{Funding}\label{funding}

The authors disclosed no financial support for the research, authorship,
and/or publication of this article.

\end{document}